%% file: main.tex
\documentclass[10pt,twocolumn,letterpaper]{article}

\usepackage[margin=0.75in,columnsep=0.25in]{geometry}

\usepackage{times}
\usepackage[utf8]{inputenc}
\usepackage[T1]{fontenc}

\usepackage{hyperref}
\usepackage{url}
\usepackage{booktabs}
\usepackage{amsfonts}
\usepackage{amsmath}
\usepackage{amssymb}
\usepackage{nicefrac}
\usepackage{microtype}
\usepackage{graphicx}
\usepackage{subcaption}
\usepackage{titlesec}
\usepackage[numbers]{natbib}
\usepackage{caption}
\usepackage{xurl}
\usepackage{enumitem}
\usepackage{placeins}

\titleformat{\section}{\normalfont\large\bfseries}{\thesection}{1em}{}
\titleformat{\subsection}{\normalfont\normalsize\bfseries}{\thesubsection}{1em}{}
\titleformat{\subsubsection}{\normalfont\normalsize\itshape}{\thesubsubsection}{1em}{}
\titlespacing*{\section}{0pt}{2ex plus 1ex minus .2ex}{1ex plus .2ex}
\titlespacing*{\subsection}{0pt}{1.5ex plus 1ex minus .2ex}{0.5ex plus .2ex}
\titlespacing*{\subsubsection}{0pt}{1ex plus 1ex minus .2ex}{0.5ex plus .2ex}

\setlist{itemsep=0.2ex,topsep=0.4ex,parsep=0pt}
\graphicspath{{media/}}

\newcommand{\email}[1]{{\small\texttt{#1}}}

\title{One Policy for 100 Household Tasks: \\
Our Solution for the 2026 BEHAVIOR Challenge}

\author{%
  \begin{tabular}{cccc}
    Ilia Larchenko & Gleb Zarin & Atindra Nair & Keerti Manney \\
    \email{ilya.larchenko@gmail.com} & \email{zaringleb@gmail.com} & \email{atindra.nair@gmail.com} & \email{keertimanney@gmail.com}
  \end{tabular} \\[1.2ex]
  \begin{tabular}{ccc}
    Vitaly Kleban & Mirko Kiefer & Anjana Yodaiken \\
    \email{vitaly.kleban@gmail.com} & \email{mirko@blackbeltlabs.com} & \email{anjana.yodaiken@gmail.com}
  \end{tabular}
}

\begin{document}

\twocolumn[
  \begin{@twocolumnfalse}
    \maketitle
    \begin{abstract}
    We describe our solution for the 2026 BEHAVIOR Challenge: 100 long-horizon household tasks in the OmniGibson simulator. We placed 9th on the final leaderboard, the only independent team in the top 10. The core of the solution is a single vision-language-action (VLA) model, one checkpoint for all 100 tasks, built on our 1st-place 2025 solution~\cite{larchenko2025behavior}, which is based on $\pi_{0.5}$~\cite{pi05_2025}. On the 50 tasks shared with 2025, its Q rises from 0.260 to 0.371; across all 100 tasks it scores 0.392.

    Our main conclusion from 2025 was that the model was undertrained, so most of the 2026 work went into training the model for longer, optimizing some training parameters, making each training step slightly faster and reducing the gradient noise:
    \begin{itemize}
        \item \textbf{migration from 50 to 100 tasks}: semantic stages built from the demonstrations' skill annotations, task and stage embeddings of new tasks warm-started from similar old ones, and a 21-dimensional action space matching the teleoperation controls;
        \item \textbf{variance-reduced multi-sample flow matching}: stratified flow times and antithetic noise, and a simple noise-vs-compute model that picks the number of action-expert samples per observation without ablations;
        \item \textbf{a 200k-step training run} (one B300, then 8$\times$H200 at batch 2048), about one pass over the 211M frames of the demonstrations;
        \item \textbf{a cloud evaluation infrastructure}: a shared evaluation record across local machines and two cloud platforms, with provenance, episode diagnostics and restorable simulator states for follow-up experiments, with six simulators on one GPU together running faster than real time.
    \end{itemize}
    On top of the learned policy we developed LLM agent control with GPT-6 Astra~\cite{openai2026astra} for 14 tasks that we routed to the agent after it outperformed the policy in our own trials:
    \begin{itemize}
        \item \textbf{an agent route}: on these tasks a GPT-6 Astra coding agent controls the robot through observation and motion tools, starting every episode from a frozen copy of its own practice session on training instances. We also explored a learned policy monitored by GPT-6.1 Sol~\cite{openai2026sol} with bounded Astra interventions, which we did not submit;
        \item \textbf{result}: across the 14 agent tasks, Astra's mean Q is 0.639, against 0.108 for the learned policy. The public submission included only part of the Astra results and scored 0.446 Q with 27.3\% full-task success.
    \end{itemize}
    \end{abstract}
    \vspace{1em}
  \end{@twocolumnfalse}
]

\section{Introduction}

\subsection{The 2026 BEHAVIOR Challenge}

The BEHAVIOR Challenge~\cite{behavior_challenge2026,li2024behavior} evaluates robot policies on long-horizon household activities in the OmniGibson simulator, built on NVIDIA Isaac Sim. The 2026 edition doubles the 2025 task set: 100 activities, from turning on a radio to cooking a meal, each with 200 teleoperated demonstrations (20{,}000 episodes, 211M frames or about 1{,}950 hours, in the LeRobot~\cite{cadene2024lerobot} format). The robot is the same Galaxea R1Pro: a holonomic base, a 4-joint torso and two 7-DOF arms with parallel grippers, observed through a head camera and two wrist cameras plus proprioception. An episode is a full success if all goal predicates are satisfied, which scores 1. Otherwise its q-score (we also call it Q) is the fraction of satisfied predicates. Scores are averaged over instances and then over tasks. The time limit of an episode is 1.5$\times$ the mean duration of the human demonstrations of the task (2$\times$ in 2025). The 50 tasks of 2025 are included with the same demonstrations; the other 50 are new.

\subsection{Vision-language-action models}

Vision-language-action models (VLAs) fine-tune a pretrained vision-language model to output robot actions~\cite{brohan2023rt2}. The $\pi_0$ family~\cite{black2024pi0,pi05_2025} pairs a PaliGemma VLM~\cite{beyer2024paligemma} with a smaller ``action expert'' trained by flow matching~\cite{lipman2023flow} to generate chunks of continuous actions, optionally co-trained with discrete FAST action tokens~\cite{pertsch2025fast}. Many alternatives have appeared since, such as GR00T~\cite{nvidia2025gr00tn1} and, increasingly prominent, MolmoAct2~\cite{fang2026molmoact2}. Yet more than a year after its release (paper in April 2025, code and weights in September 2025~\cite{openpi}), $\pi_{0.5}$ is still widely used as a base for fine-tuning: both top 2025 BEHAVIOR solutions were built on it~\cite{larchenko2025behavior,bai2025comet}. We built our 2026 solution on top of our 2025 one, so we kept its $\pi_{0.5}$-based architecture. For this reason, in parts of this report we call our learned policy simply $\pi_{0.5}$.

\subsection{LLM agents as robot controllers}

A second line of work uses large language models to control robots directly: by choosing among learned skills~\cite{ahn2022saycan}, writing control code~\cite{liang2023cap}, or orchestrating a low-level policy. With frontier models that can use tools for hours, this has become a practical option in simulation. GPT-6 Astra~\cite{openai2026astra}, released in September 2026, was strong enough to solve some BEHAVIOR tasks that our learned policy could not, so our final submission routes 14 tasks to an Astra-based agent (Section~\ref{sec:astra}).

\subsection{Our starting point}

Our 2025 solution~\cite{larchenko2025behavior} won the first BEHAVIOR Challenge. It took $\pi_{0.5}$ and changed the following: trainable task embeddings instead of text prompts, correlated noise for flow matching, a learnable mixture of all VLM layers' keys and values for the action expert, a stage-prediction head with voting (``System 2''), a FAST auxiliary loss, several flow samples per VLM forward pass, and correlation-aware inpainting at inference. The 2025 checkpoints were clearly undertrained: the loss was still falling when the deadline came, and four checkpoints split the 50 tasks between them. The LeHome Challenge solution~\cite{larchenko2026lehome} of one of our team members was built on top of the 2025 BEHAVIOR solution and introduced several engineering fixes and improvements that we inherit here as well. The 2025 runner-up~\cite{bai2025comet} also starts from $\pi_{0.5}$ and studies the effect of image resolution, of the training data and of rejection-sampling fine-tuning.

\subsection{Contributions}

\begin{itemize}
    \item A recipe for growing a task-conditioned VLA from 50 to 100 tasks without retraining from scratch: semantic stages from skill annotations, similarity-based initialization of new task and stage embeddings, and a 21-dimensional action space matching teleoperation controls (Section~\ref{sec:migration}).
    \item Registers, scratch tokens and attention sinks at several levels of a VLA; moving the pretrained SigLIP attention-sink outliers into the registers by parameter surgery on a running training made the attention maps readable (Section~\ref{sec:tokens}).
    \item Variance-reduced multi-sample flow matching and a way to choose the number of flow samples from one gradient-noise measurement (Section~\ref{sec:flow}).
    \item An evaluation infrastructure that makes cross-cloud rollouts inspectable and reusable through provenance, diagnostics and restorable simulator states, and runs six simulators per GPU faster than real time in aggregate (Section~\ref{sec:eval}).
    \item An LLM agent route with GPT-6 Astra for 14 tasks where the learned policy is weak and the agent performed better (Section~\ref{sec:astra}).
\end{itemize}

\textbf{Code and models.} Code: \url{https://github.com/Robot-Learning-Collective/behavior-1k-solution-2026}. Checkpoint: \url{https://huggingface.co/IliaLarchenko/BEHAVIOR2026VLA}. The 2025 checkpoint we start from: \url{https://huggingface.co/IliaLarchenko/behavior_submission}. Project page, including evaluation videos and visual examples: \url{https://robot-learning-collective.github.io/behavior-1k-2026/}.

\section{Overall Solution}
\label{sec:overview}

\subsection{What carries over}

From our 2025 solution we keep the model and most of the serving logic~\cite{larchenko2025behavior}:
\begin{itemize}
    \item \textbf{Task embeddings} instead of text: each task has a learned embedding, and the prompt is never tokenized.
    \item \textbf{System 2 stages}: the model predicts the current stage of the task from the base task token and is conditioned on a stage embedding; at inference the stage is tracked by voting.
    \item \textbf{Correlated flow noise}: the flow-matching noise has the covariance $\Sigma_\beta = \beta C + (1-\beta)I$ ($\beta=0.5$), where $C$ is the empirical correlation of normalized action chunks, both in training and at inference.
    \item \textbf{KV mixing}: each action-expert layer attends to a learnable mixture of the keys and values of all VLM layers.
    \item \textbf{Multi-sample flow matching}: the VLM prefix is computed once, and the action expert is run for $K$ flow samples (Section~\ref{sec:flow}).
    \item \textbf{FAST auxiliary loss}: the VLM also predicts the FAST tokens of the target chunk during training.
    \item \textbf{Correlation-aware inpainting} at inference: the actions already committed from the previous chunk anchor the next one, and the action correlations carry the correction over to the free actions.
\end{itemize}

From the LeHome solution~\cite{larchenko2026lehome} we take the training side: a weighted multi-dataset loader, corrected state scaling (in 2025 many normalized state values fell outside the range of the state bins and were clipped), loss masking of padded steps at episode ends, and the augmentation set (crop, rotation, color jitter, cutout, camera and state dropout). New in 2026, the frames after the task is complete (the idle tail of each demonstration) are never sampled for training.

\subsection{Model}

\begin{figure*}[!htb]
  \centering
  \includegraphics[width=0.92\textwidth]{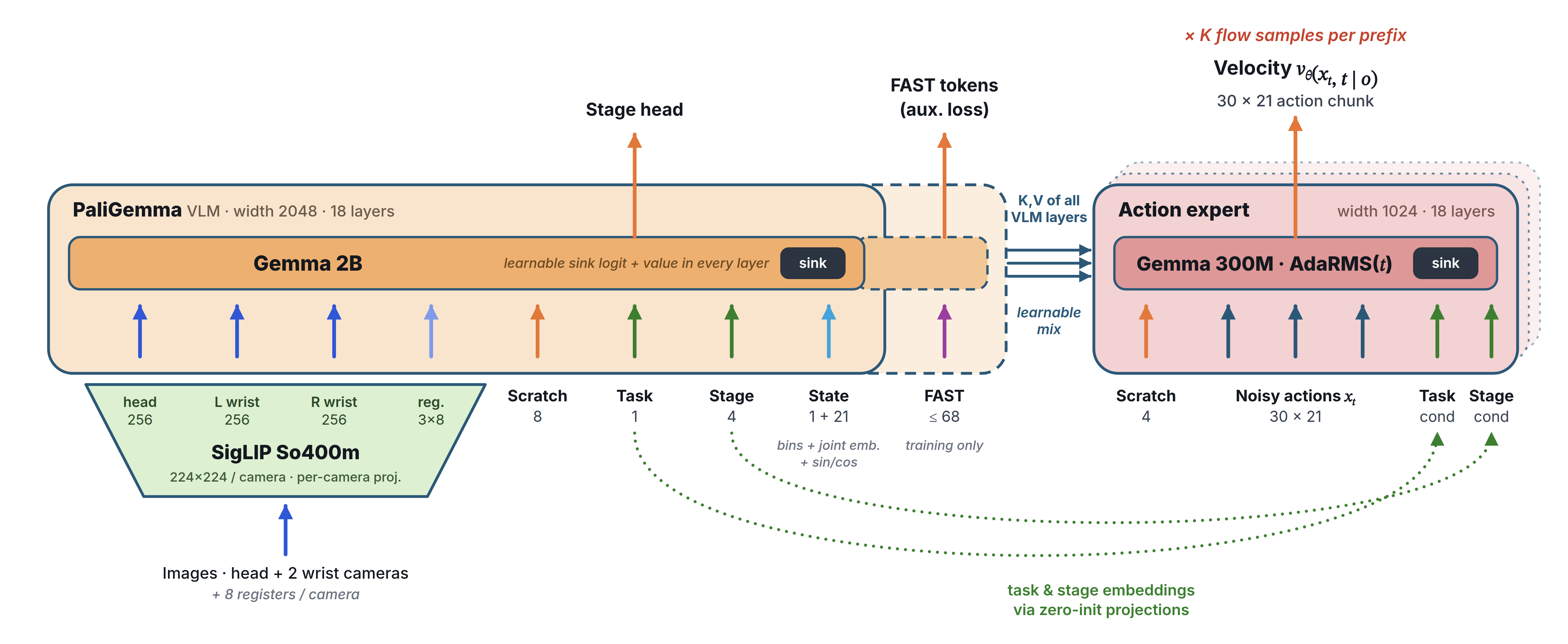}
  \caption{The 2026 model. The VLM (left) encodes three camera images, register and scratch tokens, the task and stage embeddings and the robot state; the smaller action expert (right) denoises a 30-step action chunk and reads the VLM through a learnable mixture of the keys and values of all VLM layers. The task and stage embeddings also enter the expert directly through two conditioning tokens. Every attention layer of both transformers has a learnable sink. FAST tokens are used only in training.}
  \label{fig:arch}
\end{figure*}

The backbone is that of $\pi_{0.5}$ in its openpi release~\cite{openpi}: PaliGemma (SigLIP So400m~\cite{zhai2023siglip} and Gemma 2B~\cite{gemma2024}) as the VLM and a 300M Gemma action expert with adaptive RMS-norm conditioning on the flow time (Figure~\ref{fig:arch}). Images are 224$\times$224 (256 tokens per camera). The prefix holds the image tokens with their SigLIP registers, VLM scratch tokens, the base task token, four task/stage fusion tokens, a continuous state token and 21 quantized state tokens, and in training the FAST tokens of the target chunk. The expert's suffix holds four scratch tokens, the 30 noisy actions and two conditioning tokens.

\textbf{State tokens.} $\pi_{0.5}$ embeds each state dimension as one of 256 bin tokens of the language model's embedding table, so ``left shoulder at bin 120'' and ``right wrist at bin 120'' are the same vector, and values outside $[-1,1]$ are clipped. We expanded each of the 21 state tokens to
\[
\mathrm{tok}_j = E_{\mathrm{bin}}[\mathrm{bin}(x_j)] + e_j + W\,\phi(x_j),
\]
so that the same bin maps to a distinct token for each joint, where $e_j$ is a learned per-joint embedding, $\phi$ is a 256-dimensional sine/cosine encoding of the unclipped value, and $W$ is a learned projection matrix shared across joints that maps $\phi(x_j)$ to the token width; $e_j$ and $W$ are initialized to zero. In addition to these 21 per-joint tokens, a single continuous state token is a linear projection of the whole unclipped state vector.

\textbf{Task and stage inputs to the action expert.} In $\pi_{0.5}$ the expert sees the task only through the VLM's keys and values. We add two tokens at the end of the suffix that carry zero-initialized projections of the task embedding and of the task-stage embedding, a direct conditioning path into the expert.

\begin{figure}[!htb]
  \centering
  \includegraphics[width=0.95\linewidth]{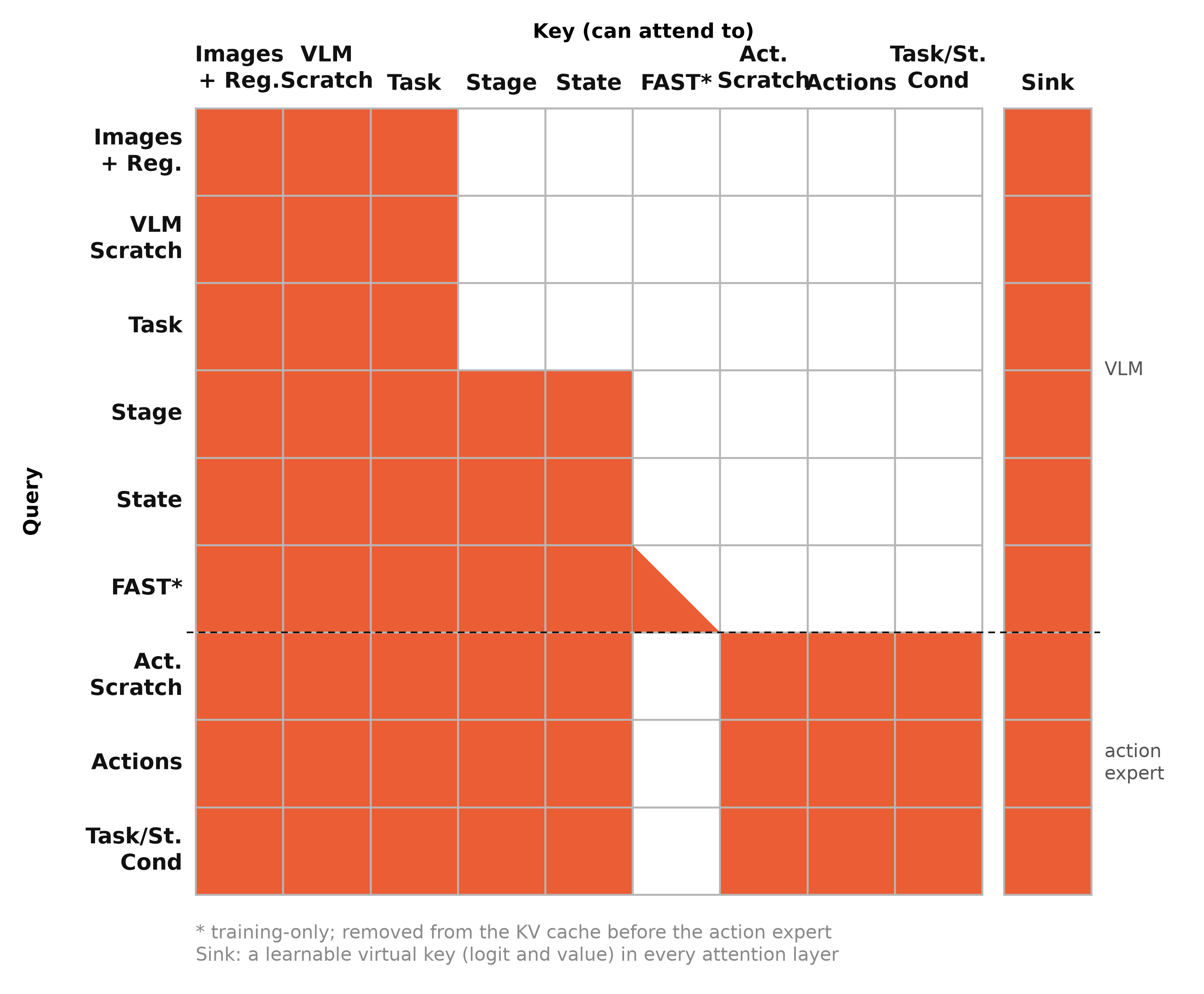}
  \caption{Group-level attention mask in training (rows: queries, columns: keys; colored: can attend). Rows above the dashed line are VLM tokens, rows below are action-expert tokens. FAST tokens are causal among themselves and are removed from the cache the expert reads. The sink column is visible to every query in every layer.}
  \label{fig:mask}
\end{figure}

\textbf{Attention structure.} Figure~\ref{fig:mask} shows the attention mask. The images, registers, VLM scratch tokens and the base task token form one bidirectional block that cannot see the stage or the state; the stage head reads the output of the base task token, so the stage is predicted from the images and the task alone. The stage and state tokens see that block and each other; the FAST tokens are causal. The action expert's tokens see the whole prefix except the FAST tokens and attend bidirectionally among themselves. Every query can also attend to the sink.

\textbf{Inference.} The serving loop is the 2025 one with re-tuned parameters; Appendix~\ref{app:inference} lists the changes.

\section{Migration from 50 to 100 Tasks}
\label{sec:migration}

The 2025 solution served four checkpoints, each fine-tuned on a subset of the 50 tasks. In 2026 one checkpoint serves all 100 tasks. It is warm-started from the 2025 checkpoint that carried the largest share of the 2025 score; a single conversion script maps it to the new task set, stage layout and action space. All parameters added in 2026 are zero-, identity- or small-normal-initialized, so the converted model starts close to the 2025 function.

\subsection{Semantic stages}
\label{sec:stages}

In 2025 the System 2 stages were equal slices of each episode in time (5--15 per task), because we did not use the skill annotations. With 20{,}000 annotated demonstrations we could check how well these time slices match the actual steps of the tasks, and they match poorly. On average only 50\% of the frames of a time stage belong to its most common annotated step (0.17--0.82 per task); 57\% of the 1{,}093 time stages of the 100 tasks have no step that covers more than half of their frames; and the borders of the time stages are no closer to the real step borders than random borders would be. Episode lengths within a task vary by 20\%, so in an 11-stage task a fixed time grid can be off by about two stages in either direction.

\begin{figure*}[!htb]
  \centering
  \includegraphics[width=0.85\textwidth]{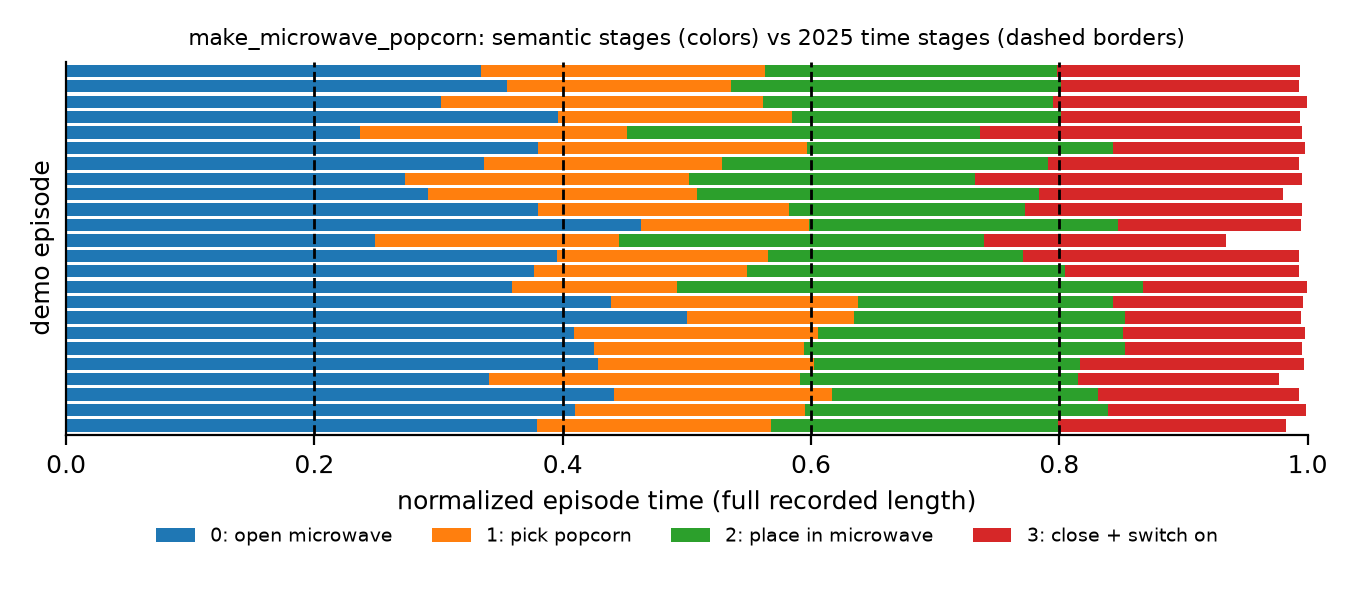}
  \caption{Semantic stages (colors) of 24 demonstrations of \texttt{make\_microwave\_popcorn} against the five equal-time stages used in 2025 (dashed). Each row is one demonstration; the white gap at the end is the idle tail after task completion.}
  \label{fig:stages}
\end{figure*}

We replaced them with semantic stages built from the annotations (Figure~\ref{fig:stages}):
\begin{enumerate}
    \item each episode is cut at the frame where the task is complete, and its sequence of annotated skills is aligned to the typical sequence of the task (the medoid); retries and extra steps are assigned to the current step;
    \item the steps are grouped into stages at points where the demonstrations agree; parts that demonstrators do in varying order (e.g.\ sorting items) become order-free progress stages;
    \item a few rules then adjust the grouping: a grasp and a press-type action (press, switch, toggle) are never in the same stage, so that the gripper rule (Appendix~\ref{app:inference}) can tell a missed grasp from a press; similar-looking states of the same appliance are kept in separate stages; repeated steps in a row become one stage; very short stages are merged into a neighbour unless they are a distinct final action.
\end{enumerate}
The result has 413 stages (2--9 per task) instead of 1{,}093 time stages. Every label is regenerated from the dataset annotations by deterministic scripts.

\subsection{Warm-starting new tasks and stages}

\textbf{Tasks.} The 50 tasks of 2025 keep their trained embeddings. We could initialize the 50 new ones randomly, but many new tasks are similar to old ones, so instead each new task starts from a weighted average of the embeddings of 1--5 similar 2025 tasks (most have three), with weights $1, 0.7, 0.5, 0.35, 0.25$ by rank. The average is rescaled to the average norm of its sources, because averaging makes vectors shorter. The similar tasks were proposed by a similarity search over task names, skills and object categories and then reviewed by hand and by Claude models. For example, \texttt{freeze\_fruit} starts from \texttt{freeze\_pies}, \texttt{clearing\_food\_from\_table\_into\_fridge} and \texttt{storing\_food}.

\textbf{Stages.} The same idea is used one level down. A semantic stage of an old task starts from the average of that task's 2025 time-stage embeddings, weighted by how many of its frames fell into each time stage. A stage of a new task starts from the matching stages of similar old tasks (e.g.\ ``open the fridge'' from the old tasks that open a fridge). The output weights of the stage head are converted in the same way.

\subsection{Minimal action space}

The 2025 model predicted 32 dimensions, of which 23 were real (base 3, torso 4, two arms of 7, two grippers). We changed the action representation to match the degrees of freedom used in teleoperation:
\begin{itemize}
    \item the 9 padding dimensions are gone;
    \item the JoyLo teleoperation system~\cite{jiang2025brs} does not move the four torso joints independently. It has two torso controls: \emph{translate} moves the torso along a fixed path through three key poses, from upright to fully bent down, and \emph{tilt} adds an offset to the third joint; the fourth joint is always zero. Every torso command in the 211M frames of the dataset has exactly this form, and translate and tilt can be recovered exactly from the four joints. The model therefore predicts translate and tilt, and its outputs are mapped back to the four joints.
\end{itemize}
The result is a 21-dim action and state space: [base 3, translate, tilt, left arm 7, left gripper, right arm 7, right gripper]. Four correlated, partly redundant torso dimensions became two meaningful ones, and the policy can no longer produce torso configurations the demonstrations never visit. The shared control space also lets a JoyLo operator smoothly take over from the policy during an episode. The weights of the remaining dimensions are copied, and the two new torso rows start at zero.

The same principle applies to the vocabulary: the model only ever embeds the 256 state bins (FAST tokens have their own embedding table), so we keep 256 of Gemma's 257{,}152 embedding rows. This removes 526M parameters that received no gradient but were stored four times in every training checkpoint (weights, EMA and the two Adam moments): about 8\,GB per checkpoint.

\section{Sinks, Scratch Tokens and Registers}
\label{sec:tokens}

\begin{figure*}[!htb]
  \centering
  \includegraphics[width=0.95\textwidth]{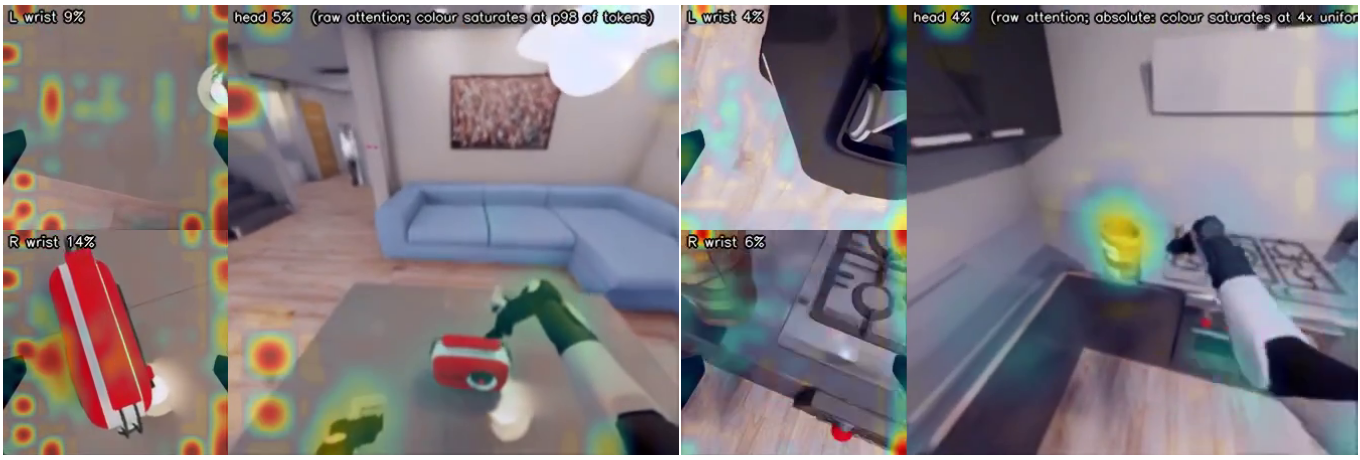}
  \caption{Raw attention of the action tokens over the three cameras (left wrist, right wrist, head). Left: the warm-started 2025 model on \texttt{turning\_on\_radio}; the same lattice of cells lights up in all cameras and frames, regardless of content. Right: the 2026 model at step 60k, after inserting registers, scratch tokens and sinks, on \texttt{make\_microwave\_popcorn}; the attention follows the popcorn bag that the robot reaches for. The color scales differ (left: 98th percentile of image tokens, right: absolute, saturating at 4$\times$ uniform).}
  \label{fig:attn}
\end{figure*}

This part of the work started as debugging. To understand failures we render the attention of the action tokens over the three cameras for every step of a rollout. For the warm-started 2025 model these maps were nearly useless: the same few cells lit up in all three cameras, in every frame of every episode (episode-mean maps correlate at 0.97, Figure~\ref{fig:attn}, left).

The cause was the pretrained SigLIP position embeddings: a few cells have outlier norms (cell (12,1) has norm 313 against a mean of 22, mostly in one channel), so their tokens act as content-free ``parking spots'' for attention. The 12 hottest cells took 31.9\% of all image attention, and cell (12,1) alone up to 29\% of the right-wrist camera. This is the artifact described by Darcet et al.~\cite{darcet2024registers} in vision transformers and by Xiao et al.~\cite{xiao2024sinks} in language models: the softmax needs somewhere to put excess attention, and the model repurposes some tokens for it, losing whatever information they carried. Beyond the lost information, it made the attention maps unreadable.

We gave the model dedicated places for this at four levels:
\begin{itemize}
    \item \textbf{SigLIP registers}~\cite{darcet2024registers}: 8 learnable tokens inside the vision tower of each camera, without position embeddings. Their outputs pass through the same projection as the patch tokens and join the prefix (24 tokens), masked together with their camera.
    \item \textbf{VLM scratch tokens}: 8 learnable tokens in the prefix, attended bidirectionally with the images.
    \item \textbf{Action-expert scratch tokens}: 4 learnable tokens at the start of the suffix.
    \item \textbf{Attention sinks} in every attention layer of both transformers: a virtual key with a learnable logit per query head~\cite{xiao2024sinks,openai2025gptoss} and, unlike gpt-oss, a learnable value per KV head, so attention parked on the sink still returns a learned vector instead of being discarded.
\end{itemize}
We also added a linear projection per camera after the shared SigLIP encoder, initialized to the identity.

\textbf{Surgery on a running training.} The new tokens start neutral in the converted 2025 checkpoint, and fresh registers did not take over the sink role by themselves: early in the production run the outlier cells were unchanged and the registers had barely moved. We therefore moved the sinks explicitly, by parameter surgery on the running training: registers were initialized from the position embeddings of the hottest cells, each such cell's position embedding was replaced by the mean of its neighbours, and the Adam moments of the edited parameters were reset. The surgery cost a brief loss bump that disappeared within a few hundred steps. After the surgery the attention maps follow the objects and the grippers (Figure~\ref{fig:attn}, right), and the attention-mass breakdown per token group became a useful diagnostic.

We did not ablate these changes. The loss fully recovered after the surgery, so we do not think they hurt, but apart from making debugging easier we have no evidence that they improve the success rate. The numbers of registers, scratch tokens and sinks were chosen fairly arbitrarily.

\section{Training Optimization}
\label{sec:flow}

\subsection{Multiple flow samples per observation}

In $\pi_0$-style training the VLM prefix is by far the most expensive part of a step, and the action expert is cheap. Our 2025 model therefore computes the prefix once and runs the expert for $K$ independent (time, noise) draws~\cite{larchenko2025behavior}. We did not ablate it then; MolmoAct2~\cite{fang2026molmoact2} later proposed the same idea and ablated it on LIBERO (average success 94.15\% with $K=1$, 95.15\% with $K=4$ and 95.90\% with $K=8$). This year we propose a simple heuristic to choose $K$ without an expensive ablation, and two ways to get more out of each sample.

Each flow sample $k$ of an observation $o$ with action chunk $a$ draws a time $t_k$ and a noise $\epsilon_k \sim \mathcal N(0,\Sigma_\beta)$, mixes them into a noisy chunk ($t=1$ is pure noise) and regresses the velocity that points from the chunk to the noise:
\[
\mathcal L(o) = \frac1K\sum_{k=1}^{K}\big\lVert v_\theta\big(t_k\epsilon_k + (1-t_k)\,a,\; t_k \mid o\big) - (\epsilon_k - a)\big\rVert^2,
\]
with $t_k\sim\mathrm{Beta}(1.5,1)$, which favors the noisy end. All $K$ samples share the prefix, and their gradients flow back into the VLM through its keys and values.

\subsection{Two sources of gradient noise}

For a batch of $B$ observations the gradient's variance has three sources: which observations are in the batch, their random augmentation, and the flow draws $(t_k,\epsilon_k)$. Only the last one shrinks with $K$:
\begin{equation}
\mathrm{Var}(B,K) = \frac1B\left(V_{oa} + \frac{V_{f}}{K}\right),\qquad V_{oa}=V_{\text{obs}}+V_{\text{aug}},
\label{eq:var}
\end{equation}
where $V_f$ is the flow variance of a single draw. We measured the three terms with a probe at fixed parameters that resamples one source at a time. In the 2025 setting ($K=15$ independent draws) the flow draws made up 21\% of the gradient variance, augmentation 38\% and the choice of observations 42\%. At our batch sizes the gradient is dominated by noise, so any reduction of its variance is worth about as much as a proportional increase of the batch~\cite{mccandlish2018largebatch}.

There are thus two ways to reduce the noise: more flow samples per observation, which shrink only the $V_f$ term, or a larger batch, which shrinks everything. Both cost compute. Below we first make each flow sample more useful (Section~\ref{sec:vr}) and then compare the two options per unit of compute (Section~\ref{sec:kopt}).

\subsection{Stratified times and antithetic noise}
\label{sec:vr}

The $K$ samples of one observation do not have to be independent; they only need the right marginal distribution, so that the expected loss and gradient are unchanged. Two classic Monte Carlo variance-reduction tools~\cite{owen2013mc} fit well:

\textbf{Stratified times.} Instead of $K$ independent draws, we take one time per quantile bin of the Beta distribution, similar to the low-discrepancy timestep sampler of variational diffusion models~\cite{kingma2021vdm}:
\begin{gather*}
t_k = F^{-1}\!\left(\frac{k + U_k}{K}\right),\qquad U_k \sim \mathcal U(0,1),\\
F^{-1}(u) = u^{2/3},
\end{gather*}
for $k=0,\dots,K-1$, independently for every observation, where $F$ is the CDF of $\mathrm{Beta}(1.5,1)$. A uniformly chosen sample is exactly $\mathrm{Beta}(1.5,1)$-distributed, so the expected loss is unchanged; what disappears is the variance between bins: a batch can no longer, by chance, put most of an observation's samples at low noise.

\textbf{Antithetic noise.} Adjacent bins share one noise draw with opposite signs~\cite{hammersley1956antithetic}: sample $2j$ uses $+\epsilon_j$ and sample $2j{+}1$ uses $-\epsilon_j$. Since $-\epsilon_j \sim \mathcal N(0,\Sigma_\beta)$ as well, the marginals are unchanged, and
\begin{gather*}
\mathrm{Var}\!\left(\tfrac12\big[f(\epsilon)+f(-\epsilon)\big]\right) = \tfrac{\sigma^2}{2}\,(1+\rho),\\
\rho = \mathrm{corr}\big(f(\epsilon), f(-\epsilon)\big),
\end{gather*}
so the component of the per-sample gradient that is odd in $\epsilon$ cancels within each pair. Pairing exact copies of $t$ with $\pm\epsilon$ was no better than pairing adjacent strata.

\begin{figure}[!htb]
  \centering
  \includegraphics[width=0.95\linewidth]{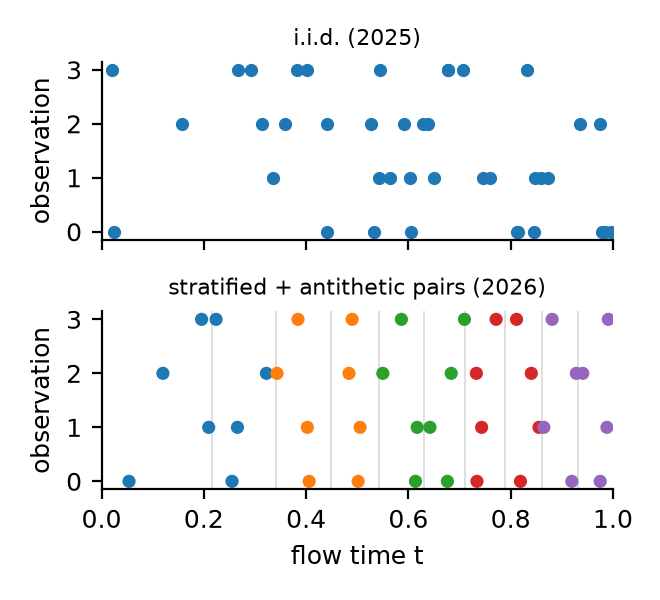}
  \caption{Flow-matching times of $K=10$ samples for four observations: independent draws (top) and stratified draws with antithetic noise pairs (bottom; gray lines are the quantile bins of $\mathrm{Beta}(1.5,1)$, colors mark the $\pm\epsilon$ pairs).}
  \label{fig:flow}
\end{figure}

Measured with the same probe, stratification halved the flow term at the same $K$ (``worth doubling $K$ for free''), and the antithetic pairs removed another 17\%. Both are a few lines of code and cost nothing.

\subsection{Choosing \texorpdfstring{$K$}{K} without ablations}
\label{sec:kopt}

\begin{figure*}[!htb]
  \centering
  \includegraphics[width=0.85\textwidth]{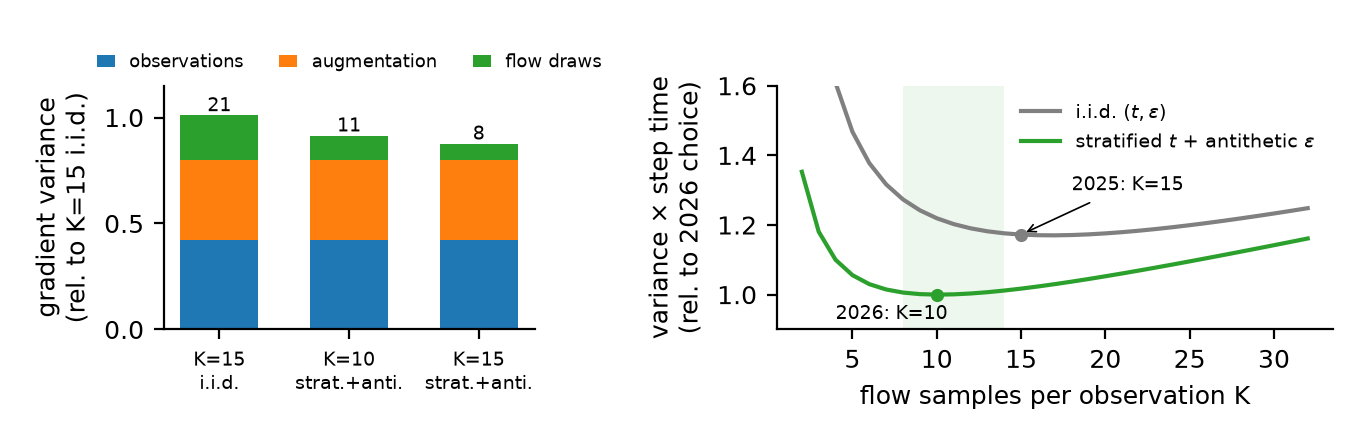}
  \caption{Left: gradient-variance budget per training step at a fixed batch, relative to the 2025 setting; only the flow term depends on the sampler and on $K$. Right: noise $\times$ step time as a function of $K$ for i.i.d.\ and variance-reduced flow samples, from the measured costs and variance ratios; the shaded band is the flat region around the optimum.}
  \label{fig:kopt}
\end{figure*}

Each extra flow sample costs compute, and so does a larger batch. With the GPU saturated, the step time is linear in both: $T(B,K) = B\,(c_p + c_k K)$, where $c_p$ is the cost of the prefix, data loading and optimizer per observation and $c_k$ the cost of one more expert pass. In the noise-dominated regime what matters is the noise per unit of compute, and $B$ drops out:
\begin{gather*}
J(K) = \mathrm{Var}\cdot T \;\propto\; (c_p + c_k K)\left(V_{oa} + \frac{V_f}{K}\right),\\
K^\ast = \sqrt{\frac{c_p}{c_k}\cdot\frac{V_f}{V_{oa}}}.
\end{gather*}
All inputs are cheap to measure: $c_p$ and $c_k$ from two 50-step runs with different $K$, and $V_f/V_{oa}$ from one probe at any $K$ (since $V_f = K\cdot V_{\text{flow}}(K)$). For the full model on one B300 we measured $c_p=21.4$\,ms and $c_k = 0.30$\,ms per observation, and $V_f/V_{oa}=1.43$ with the variance-reduced sampler, which gives $K^\ast\approx10$, with $J$ within 1.5\% of the optimum between 8 and 14 (Figure~\ref{fig:kopt}). Relative to the 2025 setting ($K=15$ i.i.d., $V_f/V_{oa}\approx3.9$), the new sampler at $K=10$ lowers variance$\times$time by about 13\%; equivalently, it is worth a 13\% larger batch for free. With the batch then set by memory (480 on one B300, 2048 on 8$\times$H200), the learning rate is scaled by the square root of the batch~\cite{malladi2022sde}.

\subsection{Other optimizations}

\textbf{Per-group learning rates.} We experimented with several per-group multipliers, applied to the final updates so that they do not touch the optimizer state and can change between resumes. The final recipe uses 2$\times$ for the action expert, the action and time projections, the KV mixing and the task/stage embeddings and 1$\times$ everywhere else, including the vision tower, with the gradient clipped at 2.0.

\textbf{Efficiency.} Capping the FAST target at 68 tokens instead of 200 (the mean is 37 per chunk) saved about 9\% of the step time, and choosing the token counts so that the attention keys are multiples of 16 another 4--5\%. Removing the unused vocabulary (Section~\ref{sec:migration}) shrinks the optimizer state and checkpoints by about 8\,GB.

\textbf{Augmentation.} Head camera: random crop (98\%) and rotation ($\pm2^\circ$); all cameras: color jitter and a cutout rectangle (probability 0.3, up to 10\% of the image); each wrist camera is dropped with probability 0.05 and all state tokens with probability 0.02.

\section{Training}
\label{sec:training}

The submitted checkpoint is step 200k of one run on all 20{,}000 demonstrations, warm-started from the converted 2025 checkpoint. The run had two phases:
\begin{itemize}
    \item \textbf{1$\times$B300}, batch 480 (the largest that fits), 130k steps, peak learning rate $2.24\cdot10^{-5}$; a step took about 10.5\,s (46 samples/s);
    \item \textbf{8$\times$H200} with FSDP, batch 2048, 70k steps, peak learning rate $4.63\cdot10^{-5}$ ($=2.24\cdot10^{-5}\sqrt{2048/480}$), continuing the same cosine schedule; a step took 11.9\,s (172 samples/s, 3.7$\times$ the B300).
\end{itemize}
The first half of the B300 phase doubled as a testbed: we tried a number of training parameters and model changes on the live run, resuming from its checkpoints, and kept what worked.  For most of the run the optimizer was AdamW with $\beta=(0.95,0.99)$ and an EMA of 0.995, the cosine schedule had 5k warmup steps, and the loss weights were 1 for flow matching, 0.1 for the stage head and 0.05 for FAST. In total the model saw about 200M training samples, roughly one pass over the 211M frames of the demonstrations.

\section{Evaluation Infrastructure}
\label{sec:eval}

Evaluating 100 long tasks is expensive: an episode runs for up to tens of minutes of wall time, and a simulator takes 6--9 minutes to boot. Over the competition we ran around 9{,}000 evaluation episodes, about 100M environment steps or nearly 1{,}000 simulated hours; we saved each run with its provenance, diagnostics and restorable scene checkpoints, so we could compare models, inspect failures and start follow-up experiments from recorded states.

\subsection{Cloud infrastructure}

\textbf{One record across clouds.} We ran evaluations on local machines and two cloud platforms (RunPod, then Nebius); all runs were uploaded to one cloud bucket. A web portal indexes them automatically. Each run carries a manifest with its full provenance: code revision and diff, simulator and package versions, all evaluator and serving arguments, and the checksum of the served checkpoint, which links the run to a model registry.

\textbf{Rich data.} Recorded evaluations retain video and resource telemetry, plus per-step actions, observation and policy diagnostics, inference timing and BDDL goal progress. We save the robot and all initialized scene objects, including sleeping and non-task objects, at the start, every 30 steps, and at termination. This gives recovery experiments a restorable scene near a failure while limiting storage and evaluation slowdown; the tradeoff is that we can restore only saved steps. When an action creates new objects, as slicing does, we also save the scene description needed to rebuild them.

\begin{figure}[!htb]
  \centering
  \includegraphics[width=0.9\linewidth]{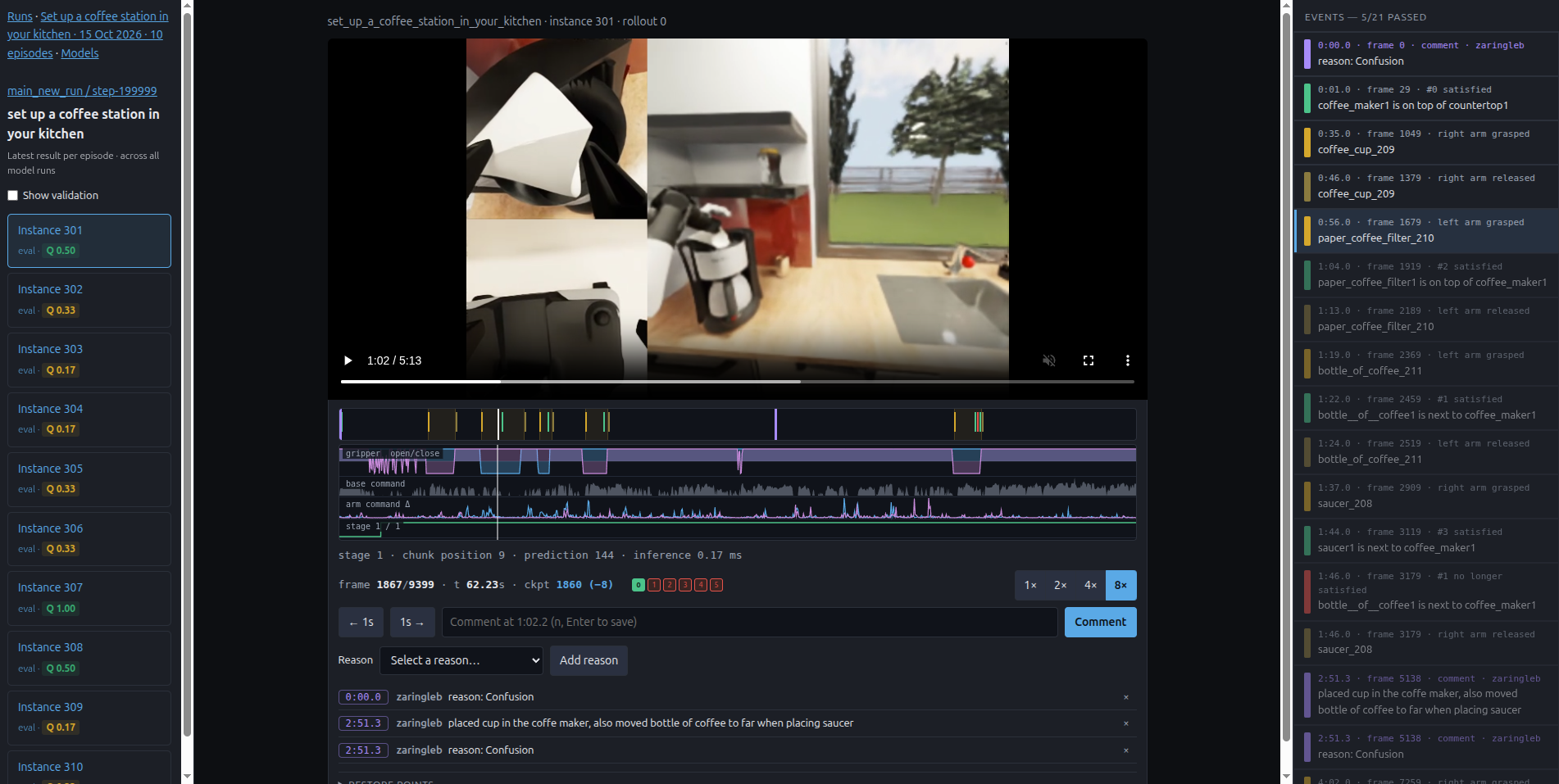}
  \caption{The evaluation portal during a coffee-station rollout: per-instance q-scores (left), synchronized three-camera video and policy traces (center), and timestamped goal, grasp, and comment events (right).}
  \label{fig:portal}
\end{figure}

\textbf{Restoration and review.} These saved states can be loaded into a fresh simulator to replay part of a rollout or start a targeted recovery experiment. The portal lets people and coding agents find any model and episode, compare outcomes, watch the video next to the goal progress and leave timestamped comments. Agent rollouts (Section~\ref{sec:astra}) use the same store and viewer (Figure~\ref{fig:portal}).

\textbf{Scaling large evaluations.} In the cloud, each GPU machine ran one policy server and six simulators, and a shared job queue, longest jobs first, kept all machines busy until the end, also when spot machines were preempted. A CPU-only head node requeued jobs after worker interruptions and uploaded completed runs. This mattered most for full evaluations: 100 tasks $\times$ 10 instances took about 20 hours on eight machines and cost about \$300. Between full evaluations, a local driver sampled the least-measured tasks for each newly published checkpoint.

\subsection{Faster than real time on one GPU}

\begin{figure}[!htb]
  \centering
  \includegraphics[width=0.9\linewidth]{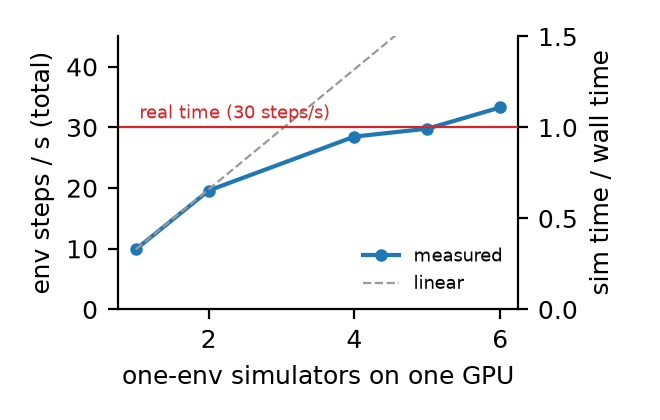}
  \caption{Throughput of one-environment simulators sharing one RTX PRO 6000 (24 vCPUs). Six simulators exceed real time.}
  \label{fig:evalscale}
\end{figure}

Simulation is CPU-bound, so the gains came mostly from keeping the CPU busy with useful work:
\begin{itemize}
    \item \textbf{Persistent simulators.} Each simulator boots once and runs many episodes. The scene and random generators are saved at boot and restored before every instance (about 10\,s instead of a 7.6-minute reboot), so each episode starts exactly like the first episode of a freshly started official evaluation. Without restoring the boot scene, changes outside the task scope---such as opened appliances, moved furniture, or created particles---can carry into the next episode.
    \item \textbf{Refilling environment slots.} With several environments per simulator, a finished slot is refilled immediately instead of waiting for the slowest episode (the official vectorized evaluator wastes 45--48\% of its steps this way); 2$\times$5 slots gave 33\% more useful steps per second than 2$\times$1.
    \item \textbf{Thread caps} for the simulator's thread pools and disabled viewport updates: a 20-episode evaluation on our 12-core workstation went from 58.4 to 33.7 minutes (1.73$\times$).
    \item \textbf{Hardware video encoding.} Videos are encoded with the GPU's NVENC encoder (H.264 for evaluation videos, HEVC for collected data) in a separate process, instead of on the CPU: at equal quality HEVC-NVENC took 2.3\,s of CPU per 300 frames vs.\ 10.4\,s for software AV1.
    \item \textbf{Native-resolution rendering.} During development the cameras were rendered directly at the model's 224$\times$224 instead of the official full resolution: about 3\,GB instead of 14--15\,GB of GPU memory per simulator and 15\% less wall time, with no measurable effect on success (28/50 vs.\ 30/50 paired episodes). The final submission renders at the official full resolution.
    \item \textbf{Batched inference.} One policy server batches all environments; a batch of 10 takes 507\,ms vs.\ 116\,ms for one.
\end{itemize}
With six simulators on one RTX PRO 6000 the machine simulates 33 environment steps per second against the 30 steps per second of simulated time: 1.11$\times$ real time (Figure~\ref{fig:evalscale}). This is an aggregate over the six simulators: each one alone runs at about 5.5 steps per second, so a single episode still takes longer than real time. Including boots and the uneven finish of a batch, a full fleet evaluation runs at about 0.86$\times$ real time per GPU.

\section{What We Tried and Dropped}
\label{sec:dropped}

\textbf{Scripted recovery in an agentic loop.} A single failed grasp ends many episodes, and the demonstrations contain almost no recoveries. We developed a small declarative DSL in which a recovery is described as a sequence of primitives (approach, grasp, rotate, release, hand back to the policy) with tolerances, and used coding agents to write recipes for selected failure modes. Recoveries are generated from full simulator-state checkpoints of failed rollouts (Section~\ref{sec:eval}), so every recovery episode starts exactly in a state the policy actually reached. On \texttt{turning\_on\_radio} the recovery data raised the specialist's success rate from about 50\% to 60--75\%, i.e.\ roughly +0.1 Q for that task; on \texttt{hanging\_pictures} it fixed the targeted failure but not the full task. Writing and validating recipes for new tasks took too long for the process to become autonomous, and at the scale of 100 tasks the expected gain for the final model was negligible.

\textbf{Manual DAgger.} We also collected human corrections~\cite{ross2011dagger} by restoring failed rollouts and teleoperating from the failure state. For \texttt{set\_\allowbreak{}up\_\allowbreak{}a\_\allowbreak{}coffee\_\allowbreak{}station\_\allowbreak{}in\_\allowbreak{}your\_\allowbreak{}kitchen}, we compared fine-tunes from the same checkpoint: one used the 200 task demonstrations, and the other sampled 70 reviewed teleop recovery segments (16{,}904 frames) at 5\% of training samples. Mean Q was 0.367 with recoveries versus 0.267 without them (one-sided paired permutation $p=0.131$), a difference that was not statistically significant.

\textbf{Astra-generated recoveries.} In exploratory experiments, we found that nearly every policy failure could, with modest manual oversight, become a 15--30\,s recovery task with a verifiable goal, such as satisfying one BDDL goal predicate or grasping or releasing an object. Astra solved all the mini-tasks we tested (100\% success) in multi-turn sessions using the RoboICL~\cite{liu2026roboicl} harness, suggesting a way to generate demonstrations for imitation learning.

Scripted recovery data and human corrections improved single tasks at small scale, and both were probably most useful for undertrained models: as the main run progressed, the base policy closed most of the gap on its own. We lacked the time and compute to scale recovery-data collection and fine-tuning to 100 tasks, or to distill Astra-generated trajectories into the policy, so we used Astra directly as a policy for difficult tasks (Section~\ref{sec:astra}).

\section{Agent Route: GPT-6 Astra}
\label{sec:astra}

\input{sections/astra}

\section{Results}
\label{sec:results}

We placed ninth overall and first among independent teams in the 2026 BEHAVIOR Challenge. On the public instances, our submission achieved a Q of 0.446 and a success rate of 27.3\%. The organizers re-evaluated only the top-ranked submissions on the private test set; ours was not among them, so our score is self-reported from our own evaluation of instances 301--310 with the official evaluator.

$\pi_{0.5}$ alone achieved a Q of 0.392. On 14 selected tasks that were hard for the policy, Astra achieved 0.639, compared with 0.108 for $\pi_{0.5}$. The public submission included only part of the Astra results due to time constraints.

\begin{figure*}[!htb]
  \centering
  \includegraphics[width=\textwidth]{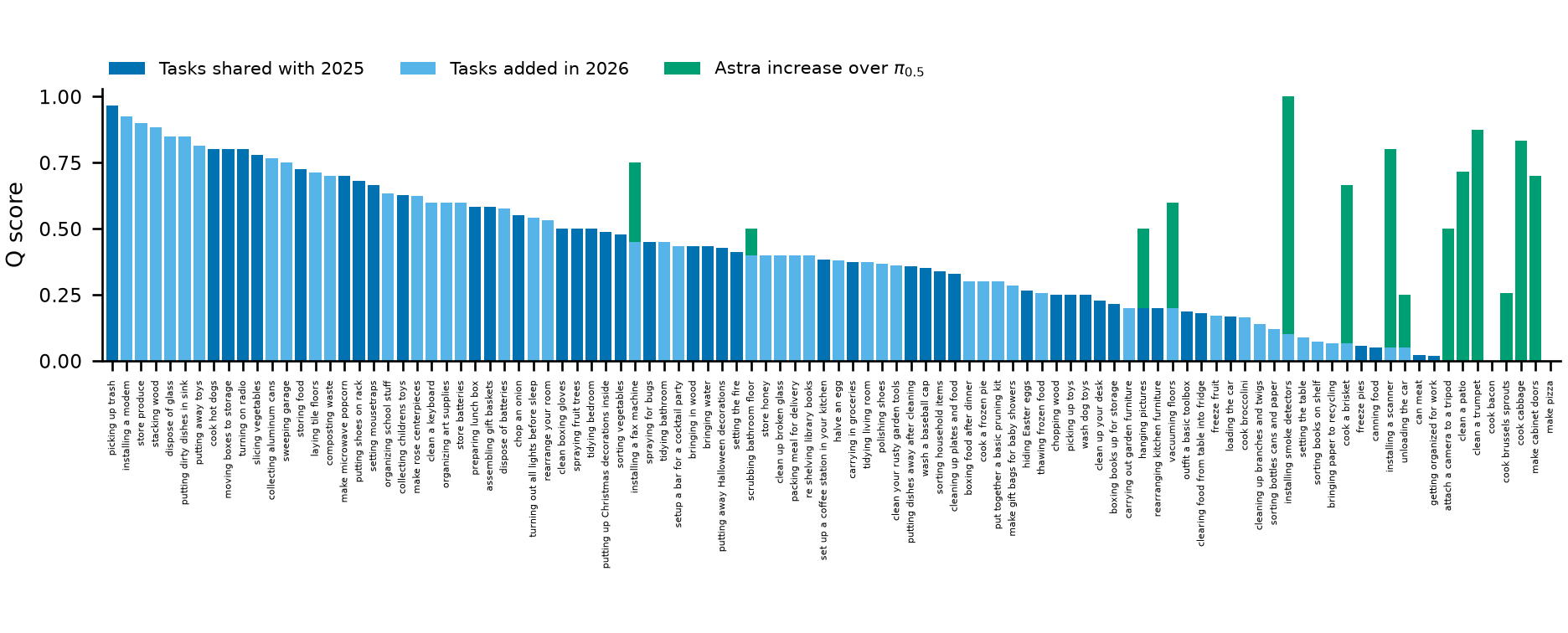}
  \caption{Public per-task Q, ordered by $\pi_{0.5}$ score. On the 14 Astra tasks, stacked bars reach Astra's Q; other bars show $\pi_{0.5}$ Q.}
  \label{fig:pertask}
\end{figure*}

The per-task $\pi_{0.5}$ scores span the full range (Figure~\ref{fig:pertask}): 13 tasks score at least 0.75 and 8 score zero.

\subsection{Comparison with 2025}

\begin{figure}[!htb]
  \centering
  \includegraphics[width=\columnwidth]{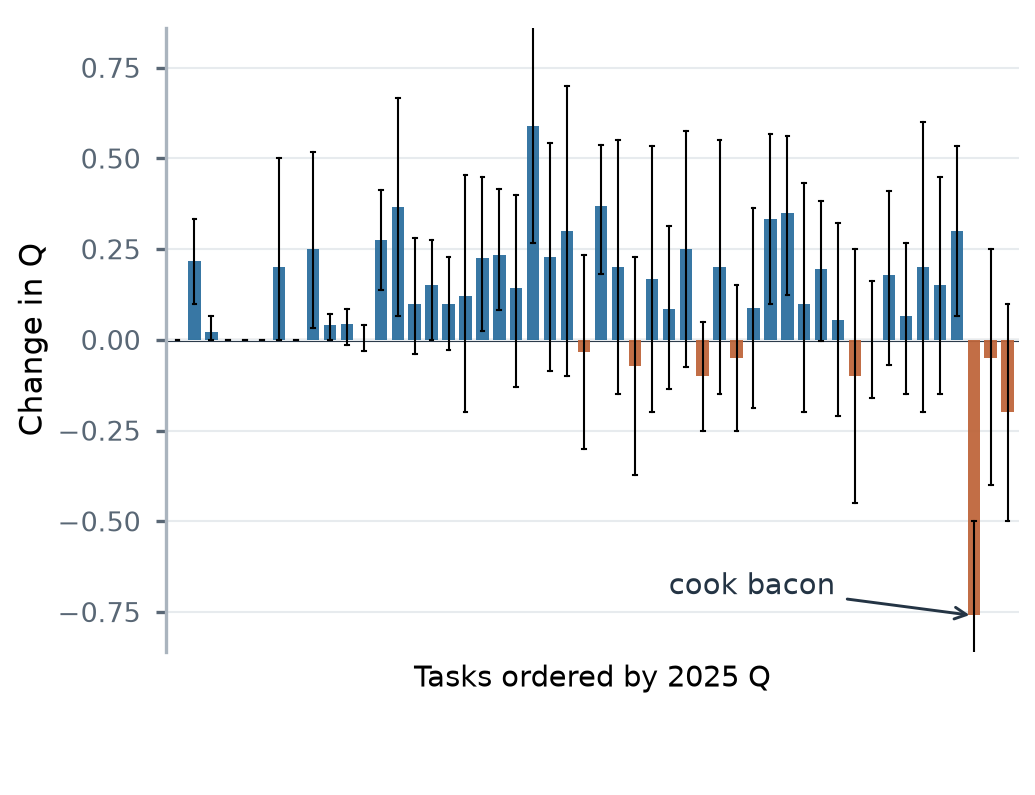}
  \caption{Change in per-task Q on the 50 tasks shared with 2025, ordered by 2025 Q (lowest at left). Bars show 2026 public-evaluation Q minus 2025 submission Q; thin lines show 95\% bootstrap intervals from ten episodes per task in each year. Missing 2025 episodes count as zero.}
  \label{fig:change-2025}
\end{figure}

On the 50 tasks from 2025, Q rose from 0.260 to 0.371 (Figure~\ref{fig:change-2025}). We think continued training, demonstrations from 50 additional tasks, model improvements, and bug fixes contributed to the gain. One checkpoint now covers all 100 tasks, compared with four checkpoints for 50 tasks last year (we did not have the resources to fine-tune multiple checkpoints). This gain came despite a 25\% shorter episode time limit (1.5$\times$ demonstration duration versus 2$\times$ in 2025). The evaluations used different simulator versions, but we did not observe a meaningful performance difference between them.

Only \texttt{cook\_bacon} was significantly worse, falling from a Q of 0.757 to 0. In the 2026 public evaluation, 7/10 instances placed the bacon tray on the lower fridge shelf (about 41 cm below its height in all 199 training demonstrations). This mismatch likely contributed to the decline, though the other three instances also scored zero.

\subsection{Comparison of failure modes}

\begin{figure*}[!htb]
  \centering
  \includegraphics[width=\textwidth]{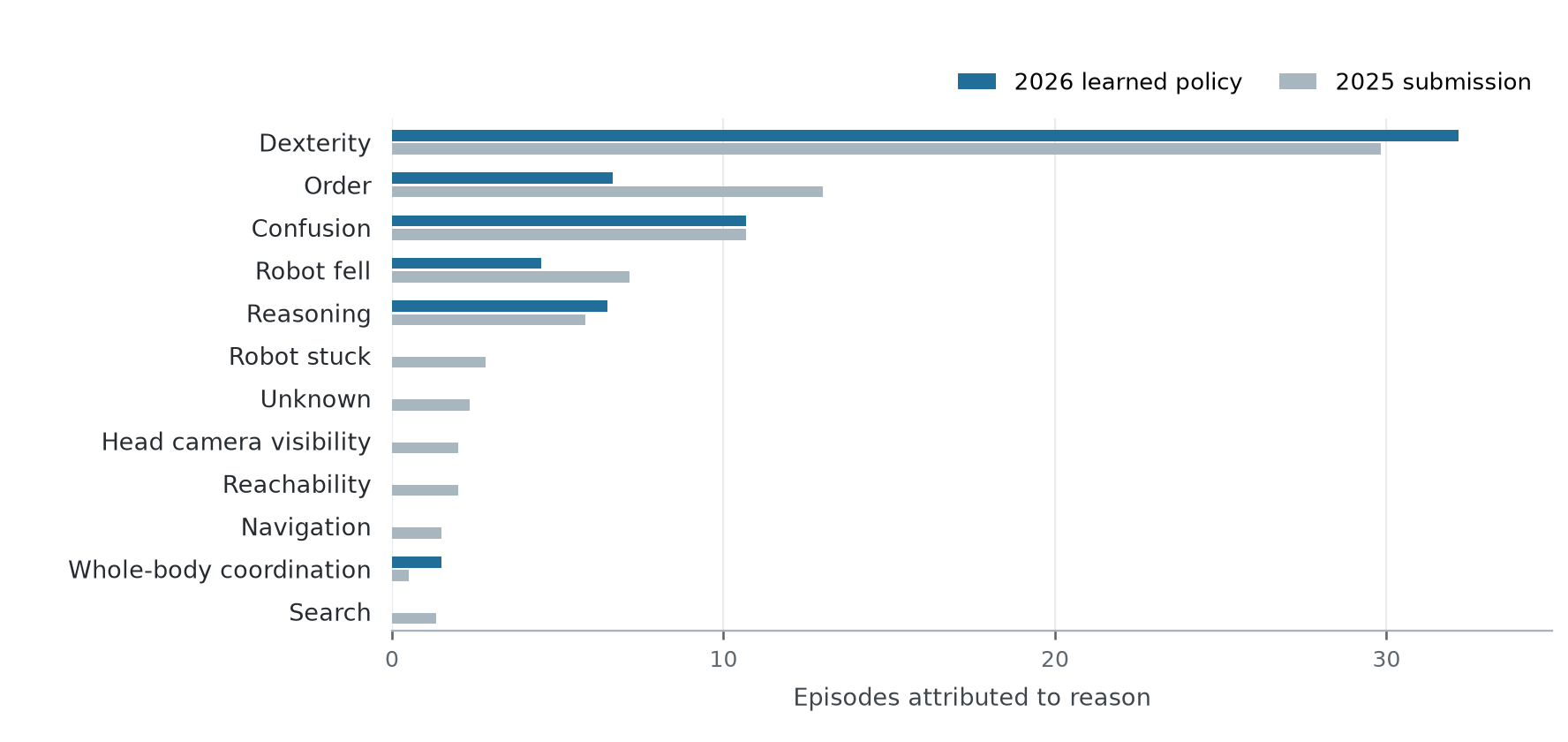}
  \caption{Manually annotated failure reasons on ten tasks shared by the 2025 submission and the 2026 learned policy. When an episode has multiple reasons, its count is divided equally among them.}
  \label{fig:failure-modes}
\end{figure*}

On ten tasks reviewed in both years, full success rose from 20\% for the 2025 submission to 36\% for the 2026 learned policy. Figure~\ref{fig:failure-modes} compares our failure-reason labels on these same tasks. The six labels assigned to 2026 failures in this sample mean:
\begin{itemize}
  \item \emph{Dexterity}: Most often, the robot cannot pick up or place objects accurately.
  \item \emph{Order}: Wrong action sequence. Demonstrations solve each task in a particular order; after an earlier step is skipped, the policy may not know how to continue (e.g., how to open a fridge when both hands already hold objects).
  \item \emph{Confusion}: Seemingly random behavior, probably after the robot enters an out-of-distribution state.
  \item \emph{Robot fell}: The robot falls, commonly after stumbling over another object.
  \item \emph{Reasoning}: A locally nonobvious action is needed (e.g., regrasping a knife by the handle instead of the blade).
  \item \emph{Whole-body coordination}: Poor coordination of arms, torso, and base when squatting, reaching, or moving through a constrained space.
\end{itemize}

The largest drop was in \emph{Order}, from 13.0 to 6.7 attributed episodes. \emph{Robot fell} decreased from 7.2 to 4.5 and \emph{Robot stuck} from 2.8 to zero, while \emph{Dexterity} remained high (29.8 versus 32.2).

Our review of the 2025 failures traced a substantial share of \emph{Order} errors to demonstration tails, where the operator kept acting after task success. These frames were included in 2025 training even though the official evaluation ends at success. Our 2026 data loader cuts demonstrations at that point, removing a training signal that often led the 2025 policy to stop before completing the goal. Better aligned semantic stages may also have reduced sequencing errors, though we did not isolate their effect. Some \emph{Order} failures remain: in \texttt{boxing\_books\_up\_for\_storage}, we verified that the stage was correct, yet the policy sometimes tried to pick up books before positioning the storage box. The policy failed to act according to the stage signal.

After switching to the 21-dimensional action representation, we observed that the robot was less likely to fall during deep squats. This representation matches the teleoperation controls, with torso translate and tilt replacing four independent torso joints, so the policy cannot command torso configurations absent from the demonstrations. The lower \emph{Robot fell} count is consistent with that observation; the same change may also explain fewer \emph{Robot stuck} failures. \emph{Dexterity} labels did not decline, although in rollout reviews we observed better recovery from manipulation failures. As fewer episodes end at earlier sequencing or stability failures, more reach difficult manipulation steps, making dexterity limitations more visible.

\subsection{Astra failure modes}

On the selected tasks we reviewed, Astra struggled mainly with precise and efficient robot control. Pouring and other dynamic movements were difficult; placement lacked precision at times; Astra occasionally knocked objects out of place or moved the robot inefficiently. It could also loop when a task appeared complete but was not. We believe limited practice time explains most failures and that more practice could improve performance substantially.

\subsection{Emergent behavior}

\begin{figure*}[!htb]
  \centering
  \begin{subfigure}[t]{0.47\textwidth}
    \centering
    \includegraphics[width=\linewidth,trim=77bp 0bp 77bp 0bp,clip]{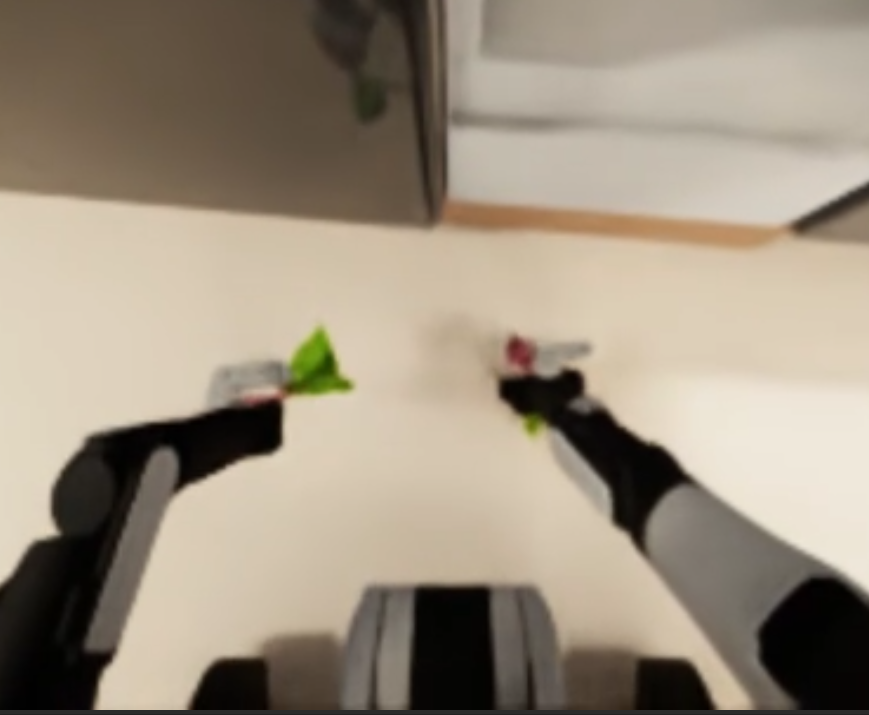}
    \caption{After dropping a beetroot, the policy efficiently picks it up and continues the task.}
  \end{subfigure}\hfill
  \begin{subfigure}[t]{0.47\textwidth}
    \centering
    \includegraphics[width=\linewidth,trim=79.5bp 0bp 79.5bp 0bp,clip]{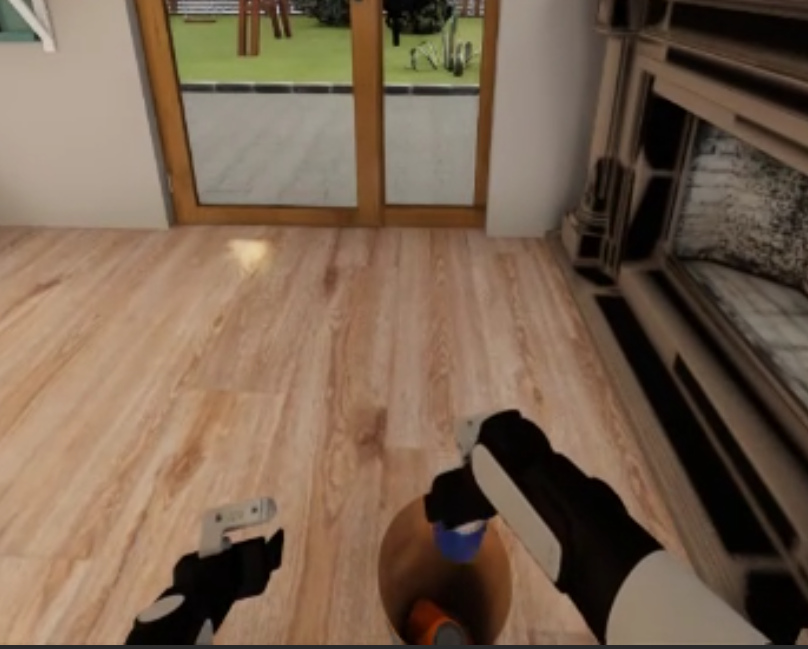}
    \caption{After putting the trash bin down too early, the policy places the can directly inside, skipping the usual step of picking up the bin.}
  \end{subfigure}
  \caption{Recovery and adaptation in the 2026 learned policy.}
  \label{fig:emergent}
\end{figure*}

Last year, we demonstrated that single-task policies made no attempt to recover, while the multi-task policy showed tentative attempts despite having no recovery demonstrations in its training data~\cite{larchenko2025behavior}. The 2026 policy, also trained without recovery examples, attempts recovery more often and with less hesitation; these attempts often let it continue the task successfully. Training across 100 tasks rather than 50 gave it a broader range of actions and situations to draw on when recovering. The policy also adapts familiar actions to the state it reaches, sometimes using the other arm or changing the action sequence after an earlier mistake (Figure~\ref{fig:emergent}). We find it promising that such goal-directed behavior can be learned from demonstrations alone across many tasks.

\section{Discussion}

\textbf{One policy on 100 tasks.} After 2025 we assumed that our model was undertrained, and this year we trained one model for 200k steps, about one pass over the 211M frames of the demonstrations. On the 50 tasks shared with 2025 it scores a Q of 0.371 against 0.260 for our winning 2025 submission (Section~\ref{sec:results}). The migration recipe (Section~\ref{sec:migration}) let us continue from the 2025 checkpoint instead of starting from scratch. One checkpoint now covers all 100 tasks (we did not have the resources to fine-tune multiple checkpoints). We cannot attribute the gain to longer training alone: the policy also saw demonstrations from 50 additional tasks, and we made model improvements and bug fixes. The 2026 episodes had a shorter time limit; the evaluations used different simulator versions, but we did not observe a meaningful performance difference between them.

\textbf{Recovery data helps a little, but we could not scale it.} Extra recovery data had a small positive effect on undertrained models: scripted recoveries raised the success rate on \texttt{turning\_on\_radio} from about 50\% to 60--75\%, and human corrections raised Q on one task from 0.267 to 0.367, a difference that was not significant. We found no efficient way to collect such data for 100 tasks, and we did not test it on the final, better-trained model.

\textbf{A practicing LLM agent solves what the policy cannot, at a high price.} After practicing on training instances in an automated research loop, GPT-6 Astra reached a mean Q of 0.639 across 14 selected tasks, compared with 0.108 for the learned policy on those tasks (Figure~\ref{fig:pertask}). On \texttt{installing\_smoke\_detectors} it scored 1.0, where the policy scores 0.10. This in-context learning from its own practice is the most striking result of our work. It is also expensive and slow: an agent episode took a median of 0.63 hours of wall time, about 17 times longer than real time, and tens of dollars in API fees, while six simulators served by the learned policy together run faster than real time on one GPU, and a full 1{,}000-episode evaluation costs about \$300. We had the agent route working only one to two days before the deadline, so it is far from tuned. Making such agents cheaper and faster, or moving what they learn into the policy, remains open.

\FloatBarrier
\section*{Contributions}

Ilia Larchenko: learned policy, general setup and training. Gleb Zarin: learned policy, evaluation infrastructure. Atindra Nair: agent control logic. Keerti Manney: evaluation optimization and agent experiments. Vitaly Kleban: final training. Mirko Kiefer: model experiments and evaluations. Anjana Yodaiken: physical JoyLo teleoperation rig and experiments. Beyond these, every team member also worked on experiments that did not make it into the final solution, some of which are described in Section~\ref{sec:dropped}.

\bibliographystyle{plainnat}

\appendix

\section{Inference Details}
\label{app:inference}

The serving loop is the 2025 one with re-tuned parameters: predict 30 actions with 20 denoising steps, execute a prefix, keep the next actions as the inpainting anchor of the next prediction, and update the stage by voting. The changes:
\begin{itemize}
    \item \textbf{Soft inpainting.} As in 2025, each denoising step moves the anchored coordinates 70\% of the way to the anchor; the correlated correction of the free coordinates ($\Sigma_{UO}\Sigma_{OO}^{-1}\delta_O$) is now capped per step instead of being dropped.
    \item \textbf{Faster playback.} The first 26 predicted actions are executed in 13 environment steps by cubic resampling, and each new chunk is shifted to start at the last executed action with an offset that decays by 0.7 per step.
    \item \textbf{Stage voting} over a fixed window of 120 environment steps, and a minimum stage duration of 0.67$\times$ the shortest duration of that stage in the demonstrations.
    \item \textbf{Gripper rule} mined from the data: a gripper may stay fully closed in a (task, side, stage) only if more than 15\% of the demonstrations close it there for at least 0.5\,s (pressing a button or switch). Elsewhere a fully closed gripper has missed its grasp, and it is opened.
\end{itemize}

\end{document}

%% file: sections/astra.tex
We used GPT-6 Astra in two ways. In the first, the learned policy controls the robot and Astra takes over only for a short recovery. In the second, Astra controls the robot for the whole episode and starts from a saved copy of its own practice on that task. Only the second is in our final submission. We describe both, in the order in which we built them.

\subsection{Sol monitoring and bounded Astra intervention}
\label{sec:astra-intervention}

In the intervention system, the learned policy stays in control. A GPT-6.1 Sol~\cite{openai2026sol} monitor watches the three camera streams at 1\,FPS and writes visual notes on the progress of the task. When the policy makes no progress for some time, for example when it fails the same grasp again and again or the robot falls, Sol writes a text description of a recovery, grounded in the task and in its notes. GPT-6 Astra then controls the robot for a bounded recovery: at most three interventions per episode, each of at most 900 actions. After the recovery, the learned policy continues from the new state.

The length of the window that Sol looks back over mattered. We started with 10 seconds. This was too short, because the failures above develop over a longer time. We increased the window to 30 seconds, which Sol reviews every 10 simulated seconds.

We found no gain that we could attribute to the interventions. On one task, two monitored runs scored a Q of 0.67, against 0.17 for the learned policy alone, but one of the two runs contained no Astra action. The monitor also costs money when it does nothing. We did not submit this system.

\subsection{Historical failure comments as context}

We also used agents to review all our earlier rollouts and to annotate where the learned policy failed. For each task we collected these comments in one file, \texttt{ANNOTATIONS.md}, and gave it to the Sol monitor, so that it knew which failures to look for and could give Astra better guidance. We had to trim the files as training progressed, because the policy changed and its early failures stopped occurring.

The comments did not change the result. With them, the intervention system scored a Q of 0 on all 8 instances that we tested of two attachment tasks: attaching a camera to a tripod and making cabinet doors. The learned policy alone also scored 0 there. These tasks give no credit for a part that is almost attached, and after a recovery the control goes back to a policy that cannot do the precise step.

This work gave us the idea for the next step. The comments were feedback that we prepared before the episode, with offline compute. Astra could prepare such feedback for itself, from its own attempts at the task.

\subsection{Practice on training instances}
\label{sec:astra-frozen}

In the final design, Astra learns each task by practice before evaluation, in the spirit of automated research loops. A practice session is one long conversation of GPT-6 Astra, running in the Pi coding agent~\cite{pi2026agent}, with one simulator. The simulator holds a training instance of the task, never an evaluation instance. It uses the same robot configuration, physics and cameras as the official evaluator, because earlier practice with a lighter robot base and different cameras did not transfer. Astra receives the task instruction, the goal definition, the human demonstrations of the task (camera images, robot state and actions) and the lessons that earlier practice sessions wrote down on other tasks.

The session follows a fixed loop. Astra first lists the hard parts of the task, such as grasps, presses, insertions, doors and hand-overs. It then practices one part at a time: it saves the scene just before the part, tries the part, and reads which goal conditions hold. When a part fails several times, Astra steps through two or three human demonstrations of the same part, compares the human motion with its own, changes what differs, restores the scene and tries again. When a part works, Astra writes a short recipe with the exact commands and measurements. After the parts, it runs the whole task from a fresh reset, without restores and within the official step limit, and writes a procedure for a later copy of itself.

Reset, save, restore and goal feedback are privileged tools that exist only in practice. No weights are trained. What Astra learns stays in its conversation, which grew to 29--395\,MB per task, and in its notes and recipes.

To test if practice transfers, we moved sessions to training instances that they had not seen, each with a different layout of the scene. The first attempt on a new instance used only the notes, without saved states, goal feedback or demonstrations. Over the final two days, 8 of 17 such first attempts succeeded. After a failed attempt, Astra studied the failure and practiced on the new instance. On \texttt{cook\_a\_brisket}, for example, the frozen session scored a Q of 0.33 and 1.0 with the official evaluator on two instances held out from practice. The learned policy scores 0.07 on this task.

\subsection{Observation and motion tools}
\label{sec:astra-tools}

\begin{figure*}[!htb]
  \centering
  \includegraphics[width=\textwidth]{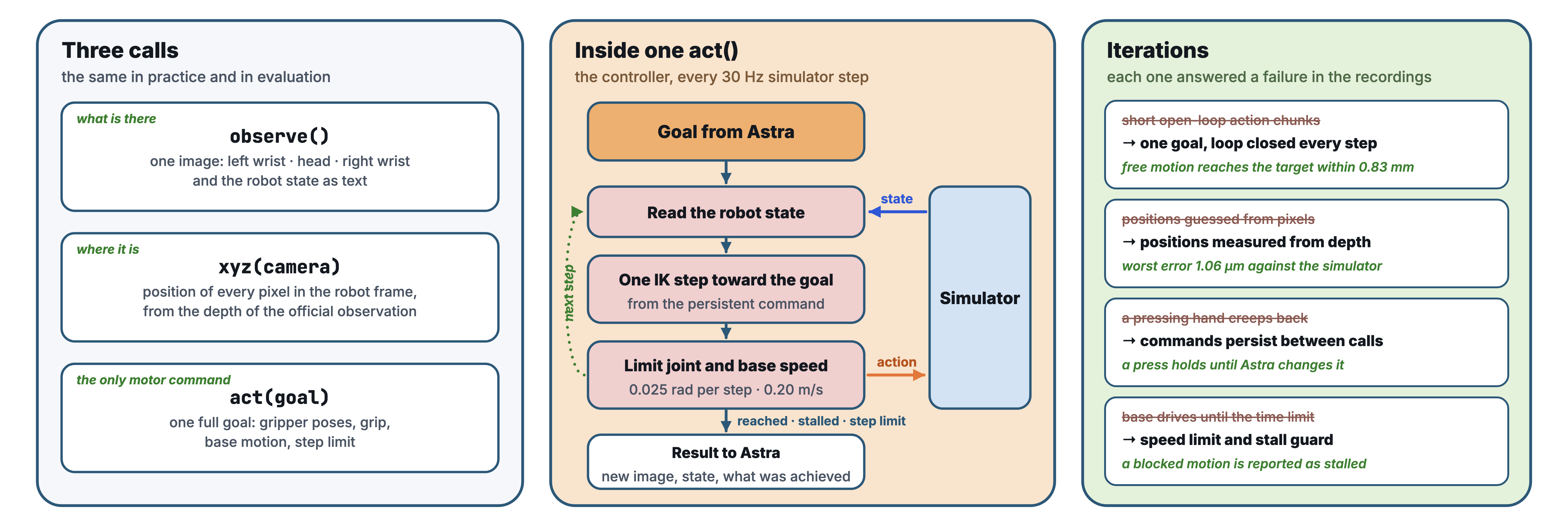}
  \caption{Observation and motion tools of the agent. Left: the three calls that Astra has, in practice and in evaluation. Middle: the controller executes one goal as one action per simulator step, and stops when the goal is reached, the motion stalls or the step limit passes. Right: the main iterations on the tools, each with the observation that confirmed it.}
  \label{fig:astra-tools}
\end{figure*}

Practice and evaluation use the same three calls and the same controller (Figure~\ref{fig:astra-tools}). \texttt{observe} returns one image with the left wrist, head and right wrist camera views, and the robot state as text. \texttt{xyz} returns the position of every pixel in the robot frame, computed from the depth image of the official observation. \texttt{act} is the only motor command: one call is one full goal, for example a gripper pose, a gripper opening and a base motion. At every simulator step (30\,Hz) a controller takes one inverse-kinematics step toward the goal and limits the joint and base speeds. It stops when the goal is reached, when the motion stalls or at a step limit, and returns a new image and what was achieved. Nothing reports contact or force, so Astra must judge from the images and the finger positions if a grasp holds.

We rebuilt these tools several times in the ten days before the deadline, and three iterations mattered most. First, we replaced short open-loop chunks of actions, after which Astra could not tell what had happened, with one goal per call and a loop that closes at every step. Free motions then reached their target within 0.83\,mm, and a blocked motion is reported as stalled. Second, we replaced positions estimated from pixels, which missed grasps by centimeters, with positions from depth. Their worst error against the simulator was 1.06\,$\mu$m over 3{,}598 samples. Third, long base motions made the robot fall, so we limited the base speed to 0.20\,m/s and stop a motion when the measured velocity does not follow the command. In our experience, these iterations mattered more than the wording of the prompts.

\subsection{The final Astra policy}
\label{sec:astra-submitted}

At the end of practice we freeze the session: we save the conversation, the notes and the recipes, and do not change them again. In evaluation, the task of an episode decides which policy controls the robot. On 14 tasks, Astra controls it from the first step to the last. On all other tasks the learned policy plays alone. Unlike in the intervention system, control does not pass back and forth.

\begin{figure*}[!htb]
  \centering
  \includegraphics[width=\textwidth]{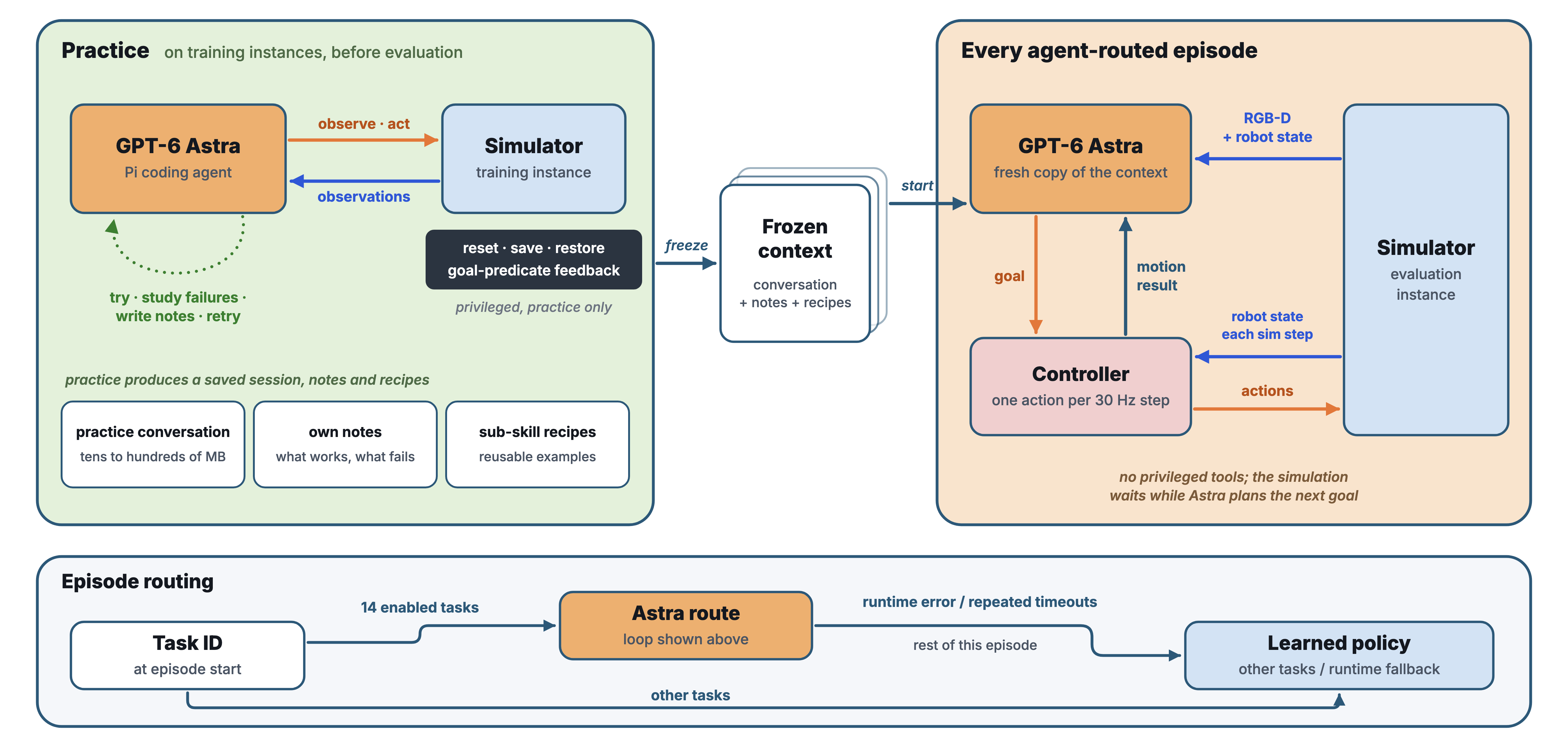}
  \caption{The Astra policy. Practice (left) uses training instances, resets, saved states and goal-predicate feedback to produce a frozen session, notes and recipes. Each agent-routed episode (right) forks that session. Astra plans goals from RGB-D and robot state; the controller uses robot-state feedback each simulator step and returns motion results, including stalls and step limits. Routing (bottom) selects Astra for 14 enabled tasks and the learned policy otherwise. Runtime errors or repeated query timeouts switch the rest of the episode to the learned policy; an unsuccessful motion alone does not trigger this fallback.}
  \label{fig:astra-policy}
\end{figure*}

Every agent episode starts from a fresh copy of the frozen session of its task (Figure~\ref{fig:astra-policy}), so nothing carries over from one episode to the next. The privileged tools are gone. Astra receives only the official observation: the head and wrist camera images with depth, and the robot state. It does not produce low-level actions. It writes short programs that call the tools of Section~\ref{sec:astra-tools}: they look at the scene, measure positions from the depth images, and send one goal at a time, such as a gripper pose, a base motion or a gripper command.  The controller converts each goal into one action per simulator step and returns the result together with a new image. While Astra thinks, the simulation waits.

If the agent fails at runtime, the learned policy plays the rest of that episode. This happens after three consecutive answers that exceed the time limit for one query, after an API error, or when the session reaches its cost limit or the episode its wall-time limit. A motion that does not reach its goal is not such a failure: Astra receives the result and plans again.

We practiced 27 tasks and routed 14 of them to the agent. These are tasks on which the learned policy is weak (mean Q of 0.108) and on which either the agent did better in our tests or the policy scores about zero, so that there was little to lose. The agent route worked only one to two days before the final deadline. Practice, task selection and tuning were therefore short, and some tasks were routed without a complete successful practice run.

Across the 14 tasks, Astra's mean Q was 0.639, against 0.108 for the learned policy alone (Figure~\ref{fig:pertask}). Due to time constraints, the public results contain only part of the agent episodes, 106 of 140. The agent is also slow and expensive: an agent episode took a median of 0.63 hours of wall time, about 17 times longer than real time, and cost tens of dollars in API fees.

